# Implementation of Radon Transformation for Electrical Impedance Tomography (EIT)

Md. Ali Hossain[1], Dr. Ahsan-Ul-Ambia[2], Md.Aktaruzzaman[3], Md. Ahaduzzaman Khan[4]

[1]Lecturer, Department of CSE, Bangladesh University, Bangladesh
ali.cse.bd@gmail.com
[2]Associate Professor, Department of CSE, Islamic University, kushtia
ambiaiu@yahoo.com
[3]Assistant Professor, Department of CSE, Islamic University, Kushtia
mazaman_iuk@yahoo.com
[4]Assistant Programmer, Bangladesh Computer Council
ahadiu19@yahoo.com

**ABSTRACT**

*Radon Transformation is generally used to construct optical image (like CT image) from the projection data in biomedical imaging. In this paper, the concept of Radon Transformation is implemented to reconstruct Electrical Impedance Topographic Image (conductivity or resistivity distribution) of a circular subject. A parallel resistance model of a subject is proposed for Electrical Impedance Topography(EIT) or Magnetic Induction Tomography(MIT). A circular subject with embedded circular objects is segmented into equal width slices from different angles. For each angle, Conductance and Conductivity of each slice is calculated and stored in an array. A back projection method is used to generate a two-dimensional image from one-dimensional projections. As a back projection method, Inverse Radon Transformation is applied on the calculated conductance and conductivity to reconstruct two dimensional images. These images are compared to the target image. In the time of image reconstruction, different filters are used and these images are compared with each other and target image.*

**KEYWORDS**

*Biomedical imaging, EIT, MIT, Radon Transformation, Inverse Radon Transformation*

## 1. INTRODUCTION

Medical imaging is the technique used to create images of the human body (or parts and function thereof) for clinical purposes (medical procedures seeking to reveal, diagnose or examine disease) or medical science (including the study of normal anatomy and physiology). Image of internal structure of a body is definitely a useful tool for diagnosis, treatment and biomedical research. Soon after the discovery of X-rays, their importance as a tool for biomedical diagnosis was recognized and X-ray machine became the first widely used electrical instrument in medicine. However, X-ray causes tissue damage in the body and cannot be used for long time monitoring [1]. In the past three decades, we have noticed tremendous developments in this field; new techniques include X-ray computed tomography (CT), magnetic resonance imaging (MRI), ultrasound imaging, positron emission tomography (PET), and so on. Each of these imaging modalities provides a particular functionality or certain unique features that cannot be replaced by the other modalities. Commonly used biomedical imaging techniques are X-ray computed tomography (CT) imaging and MRI. X-ray CT imaging has a good resolution; however, it may damage the tissue. MRI also has a good resolution and it does not cause any tissue damage. The instrument, however, is very expensive and too large in size to be used in normal patients' bedrooms in a hospital. The advantages of ultrasound imaging include bedside availability and the relative ease of performing repeated examinations. Imaging is real-time and free of harmful radiation. There are no documented side effects and discomfort is minimal [2].The disadvantages of ultrasonography are primarily related to the fact that it is heavily operator-dependent. Retrospective review of images provides only limited quality control. There is no scout scan to give a global picture for orientation [2].

Electric impedance imaging is a method of reconstructing the internal impedance distribution of a human body based on the measurements from the outside. A very low-level current requirement gives it the advantage of having no biological hazard and allows long-term monitoring for the intensive care of patients. Therefore, this method is expected to be a very efficient and useful method in the biomedical field [1].

Much research has been done and significant efforts are still being made toward the realization of electrical imaging. One common method known as electrical impedance tomography (EIT) is promising. But it involves the challenge of attaching a large number of electrodes to the body surface [3]–[5]. Another method involving magnetic excitation with coils and measurements of surface potential with electrodes has been proposed [6],[7]. In this method, however, the measurement sensitivity at the centre region is much less than that obtained in the peripheral regions. In the last decade, many studies have reported on a similar method known as magnetic induction tomography (MIT) [8]–[16]. In MIT, a magnetic field is applied from an array of excitation coils to induce eddy currents in the body, and the magnetic field from these currents is then detected using a separate set of sensing coils. The great advantage of MIT in contrast to EIT is the contactless operation, so that the positioning of the electrodes and the ill-defined electrode–skin interface can be avoided [9]. However, the sensitivity of this method is not uniform over the measurement area [9]. Another difficulty encountered with this method is that the excitation field also induces a signal in the sensing coil, and the signal due to the eddy current in the material is normally much smaller in comparison [10]. In order to overcome this difficulty, various methods have been devised [9],[11]. However, there still remain some significant drawbacks [11]. For example, using a planar gradiometer arrangement, a ghost object can be recognized in the reconstructed image [15]. In some cases, to reconstruct the image from measurement data, a weighted back projection method has been proposed [16]. Unfortunately, the weights have been calculated only in the case of conducting perturbations in empty space, which is quite different from anatomical structures [12]. In this research work a back projection method based on Inverse Radon Transformation is proposed to reconstruct 2D image of conductivity or resistivity distribution. A circular subject with embedded circular objects is taken and segmented into equal width slices from different angles. For each angle, Conductance and Conductivity of each slice is calculated and stored in an array. Inverse Radon Transformation is used to generate a two-dimensional image (Reconstructed image) from one-dimensional projections (conductivity or resistivity distribution).This Reconstructed image is compared to the target image. Different types of filters with different interpolations are used in image reconstruction process and these images are compared with each other and target image.

## 2. METHODS

In this research work, circular subjects of different sizes are taken as models. One or more small circular perturbation embedded within the circular subject as shown in Figure 1.a. The resistivity of circular subject and the embedded perturbation are known. To calculate projection of conductivity distribution at a particular angle, the subject is sliced into number of segments shown in figure 1.b where projections values for each slice is taken from $135^0$ angle. Figure1.C shows, Subject is sliced into number of segments and projections values for each slice is taken from $90^0$ angles. Average conductance and conductivity of each slice is calculated as projection value. In the similar way, the circular subject is segmented from different angles .For each angle, Conductivities and Conductance's of all slices are calculated. Thus we find the projection values of conductances and conductivities of the subject for each angle. For example, if the circular subject is divided into N slices. The circular subject is rotated by Øi angle (Øi= 1,2,3,4,.....180).Then total number of rotation is Ri =180/Øi. For each rotational angle Qc(Qc =

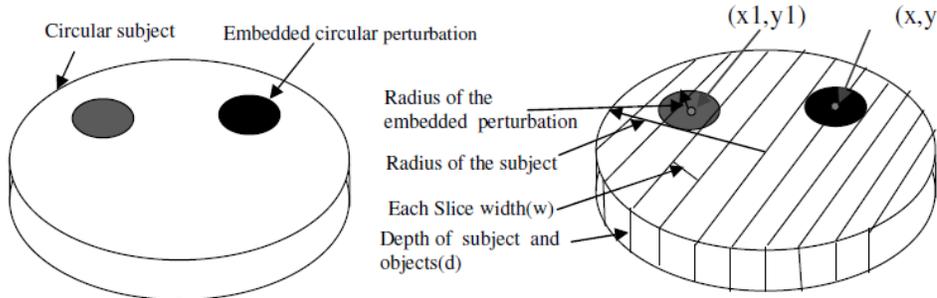

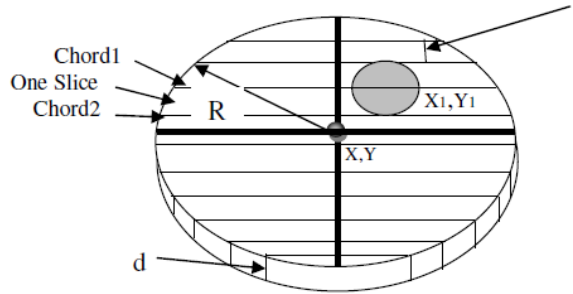

Fig1.C Segmentation of the Subject for $0^0$

Qi,Qi+Qi,......180) Conductance and Conductivity of N slices are calculated. Calculated Conductance is Conductanc[N][Ri] & Conductivity is Conductivity[N][Ri].

To calculate conductivity and conductance, each slice average length is calculated. Lengths of chords are calculated using the following formula

Chord= $2*\sqrt{Radius^2 - i^2}$ (where i = Radius-wi, w is the wide of each slice wi=0,w,2w,3w..
Radious,0,-w,-2w,-3w.....-Radious)

Each slice has two chords. Using these two chords, average length is calculated.
Using following equations, conductance and average conductivity are calculated.

We know conductance $\frac{1}{r}$ or σ = $\frac{A}{\rho l}$ ..............................................................................(1)

Conductivity $\frac{1}{\rho}$ or Average conductivity = $\frac{\sigma l}{A}$

A = average length of slice $*$ w
l is the depth of the circular subject.
ρ is the resistivity of object or subject.

If the axes of circular subject( centre fixed ) is rotated about an angle $Ø_i$, then centre of the perturbation is calculated for the new axes using following equations

$X = x1*\cos Qc + y1*\sin Qc;$ ..............................................................................................(2)

$Y = -x1*\sin Qc + y1*\cos Qc;$ ............................................................................................(3)

Where (x1,y1) is the perturbation centre for old axes and (X,Y) is the perturbation centre of that perturbation for new axes.

In each slice, one or more embedded perturbations segments can have or not.

So general equation for calculating conductance for each slice is

$$\text{Conductance} = \frac{W*Lnc}{\rho*d} + \frac{W*Lnc1}{\rho1*d} + \frac{W*Lnc2}{\rho2*d} + \ldots\ldots\ldots + \frac{W*Lncn}{\rho n*d} \quad\quad\quad\quad\quad (4)$$

Where

Lcbcesc1= chord1 – (C1+C3+…….+Cn-1) ( chord1 is one slice upper chord of the subject and c1,c3...cn-1 are chords of embedded perturbations circle1,circle2,........ circle_n . The Values of c1,c3...cn-1 can be zero or can have value depending on whether embedded perturbations chords are included with chord1 or not included with chord1)

Lcbcesc2= chord2 – (C2+C4+………+Cn) ( chord2 is one slice another chord of the subject and c2,c4...cn are chords of embedded perturbations circle1,circle2,........ circle_n . The Values of c2,c4...cn can be zero or can have value depending on whether embedded perturbations chords are included with chord1 or not included with chord2)

$Lnc = \frac{Lcbcesc1 + Lcbcesc2}{2}$     (average length of one slice of the subject without the segments of embedded circular perturbations)

$Lnc1 = \frac{C1+C2}{2}$     (average length of one slice of embedded circle1)

$Lnc2 = \frac{C3+C4}{2}$     (average length of one slice of embedded circle2)

………………..

$Lncn = \frac{Cn-1 + Cn}{2}$     (average length of one slice of embedded circle_n)

$\rho, \rho1, \rho2 \ldots \ldots \rho n$ are the resistivities of subject, embedded circular perturbations circle1, circle2,....circle_n respectively.

Conductanc[N][Ri] are One-dimensional projections data taken from different angles using conductance of each slice and Conductivity[N][Ri] are One-dimensional projections data taken from different angles using conductance of each slice.

The 2D Radon transformation is the projection of the image intensity along a radial line oriented at a specific angle.

The general equation of the Radon transformation is acquired [17]–[21]

$$g(s,\theta) = \iint f(x,y) \cdot \delta(x\cos\theta + y\sin\theta - s)\,dxdy \quad\quad\quad\quad\quad (5)$$

To reconstruct the image from one-dimensional projections data from different angles, the Inverse Radon Transform is applied to the one-dimensional projections data. The inverse of Radon transform is calculated by the following equation [22] :

$$f(x, y) = \int_{-\pi/2}^{\pi/2} \rho \cdot R_\theta(s(x, y)) d\theta \quad \ldots (6)$$

where $R_\theta$ is the Radon transformation, $\rho$ is a filter and $s(x, y) = x\cos\theta + y\sin\theta$

To get f(x,y) back, from equation (5) is known as inverse Radon transform. Figure 2.b shows the formation of image from the projections taken at angles 0°, 45°, and 90° that is shown in Figure 2.a.

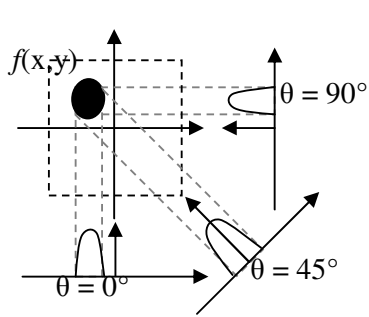
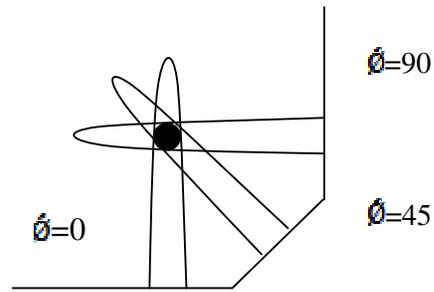

Figure 2.a  Projection from 2D image

Figure 2.b Image reconstruction from projections data

Using Inverse Radon transformation on conductivity[N][Ri], coductance[N][Ri] images are reconstructed. Hann,Sheep-Logan, Hamming,cosine and Ram-Lak filters with linear, spline and nearest interpolations are used to produce better images. These images are compared to the target image that is considered as a base image.

## 3. EXPERIMENT AND RESULTS:

### One embedded circular perturbation in the Subject:

We took Radius of the subject, R=40mm, embedded object r1=10mm,resistivity of the subject, p=0.0005 Ωm, resistivity of the object p1=0.0002 Ωm, Depth of the subject d=2mm and slice wide w=1mm, rotational angle q=$10^0$ & centre of the embedded object (x,y)=(10,10), then expected image is shown in Figure 3. slice Vs Conductance and slice Vs average conductivity are shown in Figure 4.a and Figure4.b for $0^0$ angle respectively and Figure 4.c & Figure 4.d for 0,5,10….180 degree angles respectively.

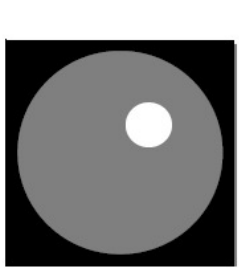

Figure 3 Expected image

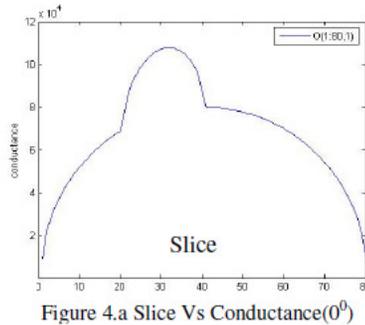

Figure 4.a Slice Vs Conductance($0^0$)

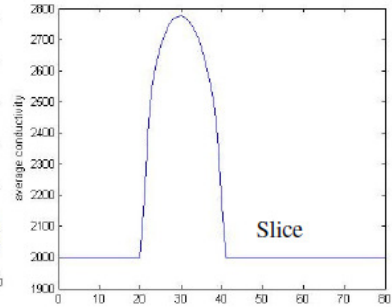

Figure 4.b Slice Vs average conductivity ($0^0$)

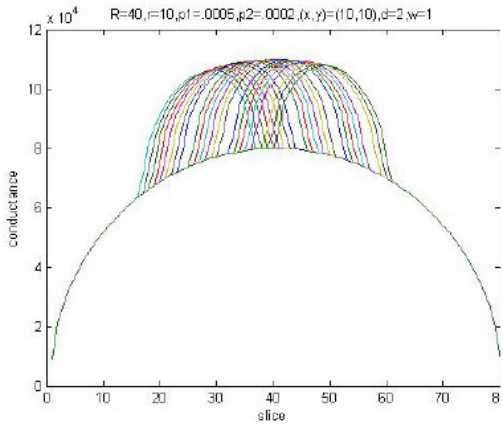

Figure 4.c Slice Vs Conductance($0^0, 5^0, 10.......180^0$)

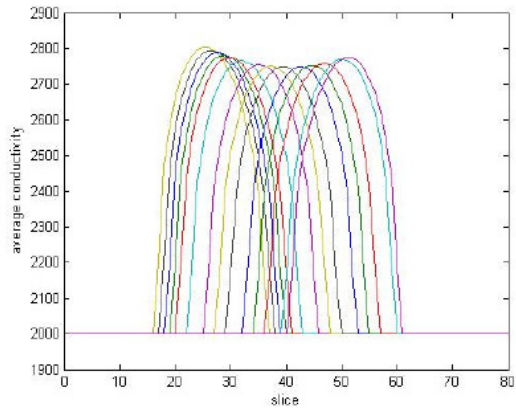

Figure 4.d Slice Vs average conductivity ($0^0 5^0, 10...180^0$)

Figure 4 shows Slice Vs Conductance & Slice Vs average conductivity graph for different angles. Using Inverse Radon Transformation on the calculated average Conductivity and conductance for all angles image is obtained Shown in Fig 5.a & 5.b.

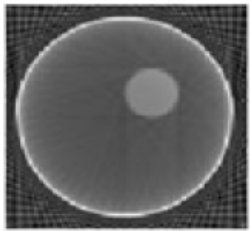

Fig 5.a Normalize average conductivity image

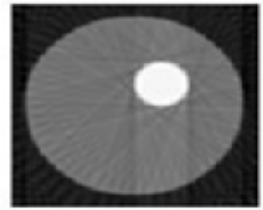

Fig 5.b Normalize conductance image

Figure 6 shows normalize average conductivity images, normalize conductance images & target images for R=40mm,r1=10mm,p1=.0005Ωm,p2=.0002Ωm,(x,y) = (-10,10), w=1mm,d=2mm, q=10&R=40mm,r1=10mm,p1=.0005Ωm,p2=.0002Ωm,x=10,y=-10,w=1mm,d=2mm,q=10

**Two embedded perturbations in the Subject:**

we took Radius of the subject R=40 mm, objects(**perturbation**) r1=8mm, r2=12mm, resistivity of subject p1=0.0005 Ωm, resistivity of embedded objects p2=0.0002 Ωm,p3=.000199 Ωm,

Depth of subject d=2mm and slice wide w=1mm, rotational angle q=5 and centre of the embedded objects (x,y)=(10,14),(x1,y1)=(-12,-10) , the expected image is shown in Figure 7. Slice Vs Conductance and slice Vs average Conductivity are shown in Figure8.a. and Figure 8.b respectively for 0 (zero) degree angle and Figure 8.c & Figure 8.d for 0,5,10….180 degree angles respectively.

Using Inverse Radon Transformation on the 1D calculated average Conductivity and conductance for all angles($0^0$ ,$5^0$,10…….$180^0$) images are obtained as shown in Figure 9.

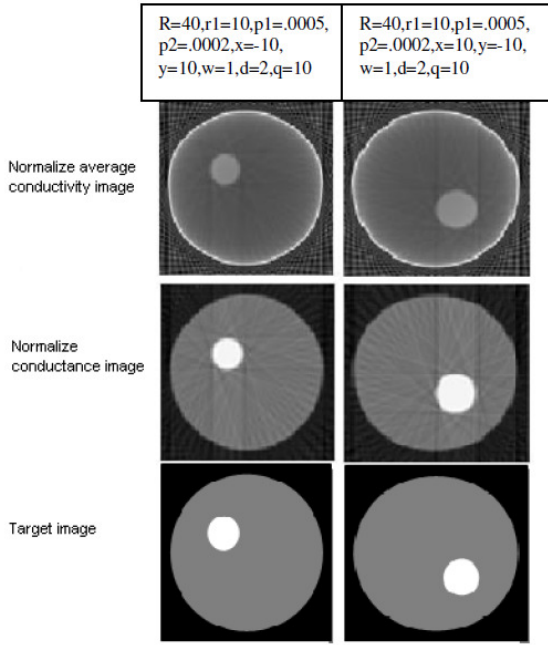

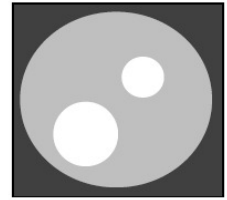

Figure 7 Expected Image

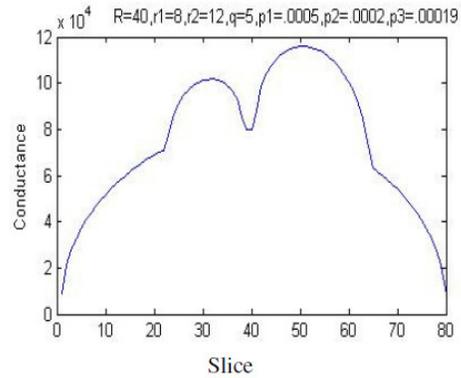

Figure 6. Normalize average conductivity , normalize conductance & target images for different positions of embedded circle

Figure 8.a Slice Vs Conductance ($0^0$)

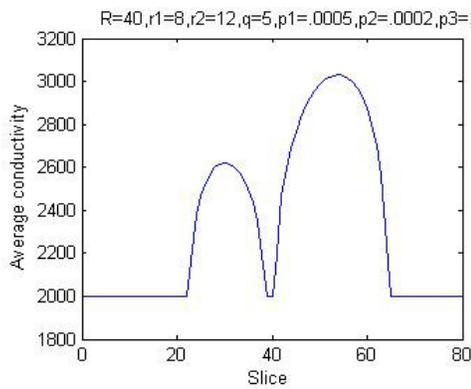

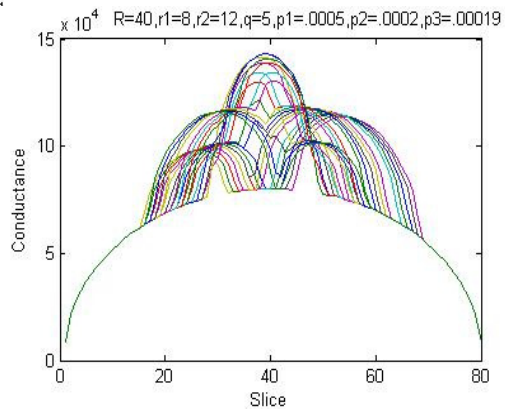

Figure 8.b Slice Vs average conductivity ($0^0$)    Figure 8.c Slice Vs conductance  ($0^0$ ,$5^0$,10…$180^0$)

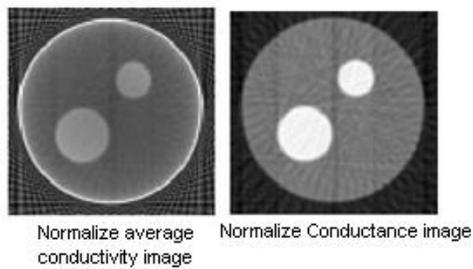

Figure 9. Normalize average conductivity & conductance images, where R=40mm, r1=8mm, r2=12mm , p1=0.0005 Ωm, p2=0.0002 Ωm p3=.000199 Ωm, ( x,y)=(10,14), (x1,y1)=(-12,-10)

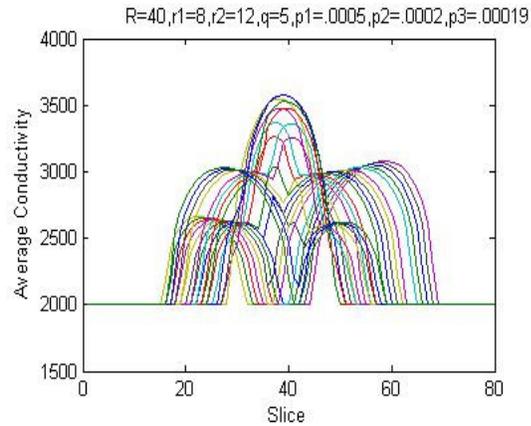

Figure 8.d SliceVs average conductivity ($0^0$,$5^0$,10…$180^0$)

We used radius and resistivity of the subject R = 40mm   p1=.0005 Ωm, for embedded perturbations, radius are  r1=10mm, r2=5mm, & resistivity p2=.008 Ωm,p3=.009 Ωm & centre positions are (x,y)=(10,10), (x1,y1)=(-8,8) respectively, slice width w=1mm, Depth of the subject and  objects is d=2, rotational angle  q=5, then obtained normalize and without normalize average conductivity and conductance images without  any filters is shown in Figure-10.

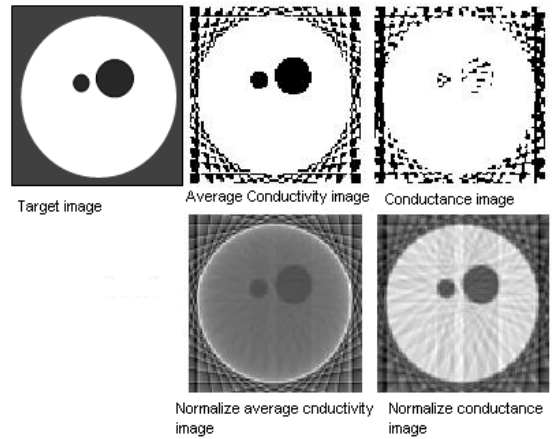

Figure 10.Target image, normalize and without normalize  average conductivity and conductance images, where R=40mm,r1=10mm,r2=5mm, p1=0.0005Ωm ,p2=0.008Ωm  p3=.009 Ωm, (x ,y)=(10,10),  (x1,y1)=(-8,8)

At the time of producing image from 1D projection data by the Inverse Radon Transformation, Interpolations (linear,nearest,splin) and  filters (Hann , Shepp-Logan, Cosine, Hamming & Ram-Lak) are used to produce different images. Figure 11 shows average conductivity images   for spline_Ram-Lak ,spline _Cosine, nearest_Ram- Lak, nearest _Hamming, nearest_Cosine, nearest _Shepp-Logan,     nearest     _Hann     (Iterpolation_filter)where     R=40mm r1=10mm,r2=5mm,p1=0.0005Ωm,     p2=0.008Ωm,p3=.009Ωm,     q=5,     (x1,y1)=(-8,8) &(x,y)=(10,10)

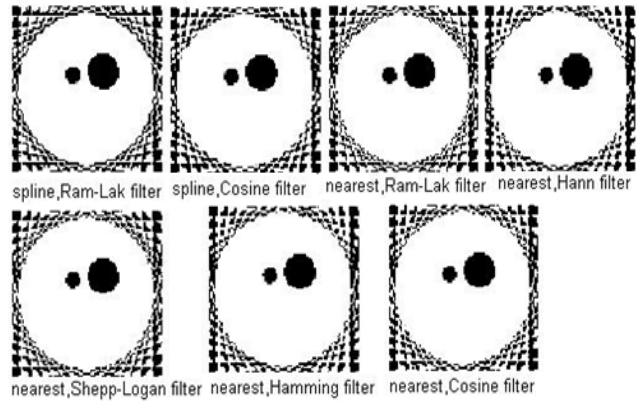

Figure 11.average conductivity images for spline_Ram-Lak ,spline _Cosine, nearest_Ram- Lak, nearest _Hamming, nearest_Cosine nearest _Shepp-Logan, nearest _Hann (Iterpolation_filter)

Figure12. shows conductance images for spline- Ram-Lak spline-Cosine, nearest-Ram-Lak, nearest-Hann nearest-Shepp-Logan, nearest-Hamming, nearest-Cosine (Interpolation-Filter). Where R=40mm,r1=8mm,r2=14mm p1=.000001Ωm, p2=.05 Ωm, p3 = .09 Ωm, (x,y)=(14,12) (x1,y1)=(-14,-10), w=1mm, d=2mm, q=$5^0$

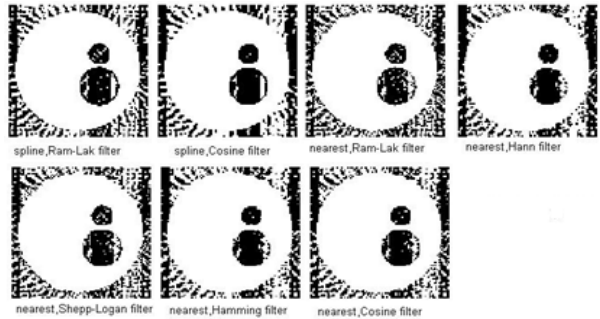

Figure 12.Conductance images for spline_Ram-Lak ,spline _Cosine, nearest_Ram- Lak, nearest _Hamming, nearest_Cosine nearest _Shepp-Logan, nearest _Hann (Interpolation_filter)

**Three embedded perturbation in the subject:**

we took the Radius of the subject R=40mm, objects(embedded circle) r1=10mm, r2=12mm, r3=8mm and resistivity for the subject is p1=0.0005Ωm and resistivities for objects are p2=.0999Ωm, p3=.0699Ωm,p4=.0699Ωm& centre position of the embedded circles are (x,y)=(10,15), (x1,y1)=(-10,15), (x,y)=(-10,-15) respectively, Depth of subject and objects d=2mm,rotational angle q=$5^0$, then obtained normalize and without normalize average conductivity and conductance images without any filters is shown in Figure 13.

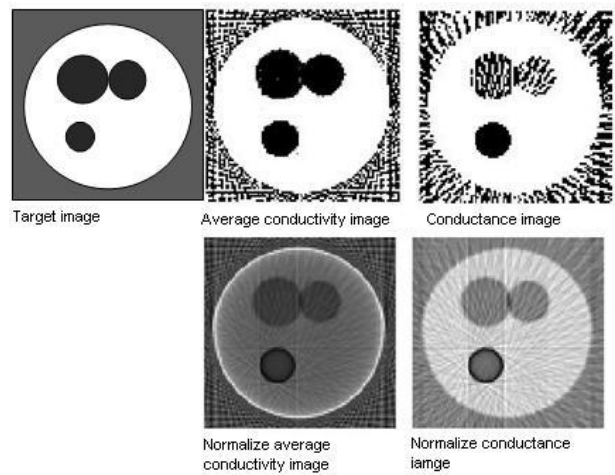

Figure 13.Target image, normalize and without normalize average conductivity and conductance images, where R=40mm r1=10mm r2=12mm ,r3=8mm ,p1=0.0005 Ωm, p2=.0999,p3=.0699,p4=.0699 (x,y)=(10,15),(x1,y1)=(-10,15)  (x,y)=(-10,-15)

## 4. DISCUSSION AND CONCLUSION

In this thesis work, we have proposed a parallel resistance model for Electrical Impedance Imaging. With some simulation study, we have shown that this model can be used to reconstruct internal resistivity or conductivity or conductance distribution images. There are some differences in target image and reconstructed image. Different types of filters with different interpolations are used in image reconstruction process. But in the reconstructed images, no significant difference is observed. If the rotational angle is small then produced images are better. From the reconstructed images with different condition we can conclude that this model can be used for Electrical Impedance Tomography (EIT) or Magnetic Induction Tomography(MIT).

## Authors

**Md. Ali Hossain** was born in Manikganj, Bangladesh, on 30 December 1984. He received the B.Sc. and M.Sc. degrees from the Department of Computer Science and Engineering, University of Islamic University, Kushtia, Bangladesh, in 2008 and 2009, respectively.He is serving as a Lecturer with the Department of Computer Science and Engineering (CSE), Bangladesh University, Dhaka. His current research interests include biomedical imaging, biomedical signal and speech processing and bioinformatics, AI.Mr. Ali Hossain is an Associate Member of the Bangladesh Computer Society and Executive Member of Islamic University Computer Association (IUCA).

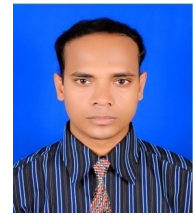

**Ahsan-Ul-Ambia** was born in Nawabgonj, Bangladesh. He received the B.Sc. and M.Sc. degrees from the Department of Applied Physics and Electronics, University of Rajshahi, Rajshahi, Bangladesh, in 1996 , and 1998 respectively. He received the PhD degree from the Graduate School of Science and Technology, Dept. of Information Science and Technology, Shizuoka University, Japan, in September 2009 respectively. He was an Assistant Programmer in a private company. He is serving as an Associate Professor with the Department of Computer Science and Engineering (CSE), Islamic University, Kushtia. He joined at Islamic University in October 1999 after completion his M.Sc. degree. His current research interests include biomedical imaging, biomedical signal and image processing, and

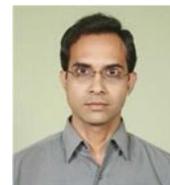

International Journal of Information Sciences and Techniques (IJIST) Vol.2, No.5, September 2012

bioinformatics. He has several papers related to these areas published in national and international journals as well as in referred conference proceedings. Dr. Ambia is an Associate Member of the Bangladesh Computer Society and a member of the Institute of Electronics, Information and Communication Engineers, Japan.

Md. Aktaruzzaman was born in kushtia, Bangladesh. He received the B.Sc. and M.Sc. degrees from the Department of Computer Science & Engineering, Islamic University, Kushtia, Bangladsh. He is serving as an Assistant Professor with the Department of Computer Science and Engineering (CSE), Islamic University, Kushtia. His current research interests include Medical Image Processing, Image Processing and Bioacoustics.. He has a number of papers on Image and Speech Processing, published in some national and international journals . 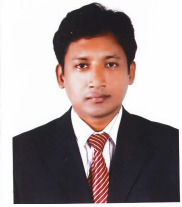

**Md. Ahaduzzaman Khan** was born in Faridpur, Bangladesh, on December 9,1987. He received the B.Sc. and M.Sc. degrees from the Department of Computer Science and Engineering, University of Islamic University, Kushtia, Bangladesh, in 2008 and 2009, respectively. He is serving as an Assistant Programmer in Bangladesh Computer Council. His current research interests include biomedical imaging, biomedical signal and image processing, and bioinformatics. Mr. Khan is an Associate Member of the Bangladesh Computer Society. 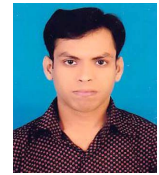